\pdfoutput=1

\documentclass[11pt]{article}

\usepackage[]{acl}
\usepackage{fontawesome5}

\usepackage{times}
\usepackage{latexsym}
\usepackage{multirow}
\usepackage[normalem]{ulem}
\usepackage[T1]{fontenc}
\usepackage{pythonhighlight}
\usepackage{amsmath}
\usepackage{verbatim}
\usepackage{paralist}
\usepackage{framed}
\usepackage{xcolor}
\usepackage{graphicx}
\usepackage{subcaption}
\usepackage[most]{tcolorbox}
\usepackage[utf8]{inputenc}
\usepackage{enumitem}
\usepackage{CJKutf8}
\usepackage{microtype}

\usepackage{inconsolata}
\usepackage{soul}
\usepackage{amssymb}
\usepackage{booktabs}
\usepackage{graphicx}
\usepackage[export]{adjustbox}
\usepackage{soul}
\usepackage{xcolor}
\usepackage{soulpos} 
\usepackage{etoolbox}
\usepackage{colortbl}
\usepackage{siunitx}
\usepackage{tabularx}
\usepackage{icomma}
\usepackage{pifont}

\usepackage{pgfplots}
\pgfplotsset{compat=1.17}
\usepgfplotslibrary{groupplots}
\usepackage{subcaption}
\usepackage{siunitx}
\definecolor{metricS}{HTML}{1f77b4} 
\definecolor{metricP}{HTML}{ff7f0e} 
\definecolor{metricE}{HTML}{8c564b} 
\definecolor{findinggreen}{HTML}{E8F9E8}
\usepackage{algorithm}
\usepackage{algorithmicx}
\usepackage{algpseudocode}
\usepackage{amsfonts}  

\usepackage{graphicx}
\usepackage{booktabs}
\usepackage{makecell}
\definecolor{myorange}{HTML}{dd614e}
\definecolor{myyellow}{HTML}{876e2f}
\usepackage{array, booktabs, ragged2e}
\newcolumntype{R}[1]{>{\RaggedLeft\arraybackslash}p{#1}}
\newcolumntype{L}[1]{>{\RaggedRight\arraybackslash}p{#1}}

\definecolor{mwpblue}{HTML}{0171C8}
\definecolor{mwpred}{HTML}{EF5544}

\usepackage{xcolor}
\definecolor{rs2}{HTML}{1f77b4}
\definecolor{rs3}{HTML}{2ca02c}
\definecolor{rs4}{HTML}{ff7f0e}
\definecolor{rs5}{HTML}{d62728}

\newcommand{\rslegend}{
  \textcolor{rs2}{\textbf{blue}} for $rs=2$, 
  \textcolor{rs3}{\textbf{green}} for $rs=3$, 
  \textcolor{rs4}{\textbf{orange}} for $rs=4$, and 
  \textcolor{rs5}{\textbf{red}} for $rs=5$
}

\newcommand{\dataset}{\textit{GSM-DC}}

\usepackage{amsthm}
\theoremstyle{definition}

\newtcolorbox{findingbox}[1][]{
    breakable,
    enhanced,
    sharp corners,
    boxrule=0pt,
    colback=findinggreen,
    colframe=findinggreen,
    frame hidden,
    borderline west={2pt}{0pt}{green!70!black},
    left=6pt,
    right=6pt,
    top=4pt,
    bottom=4pt,
    before skip=10pt,
    after skip=10pt,
    fontupper=\linespread{1.0}\selectfont,
    #1
}

\newcommand{\researchfinding}[1]{%
  \begin{findingbox}
    \faLightbulb[regular]~#1
  \end{findingbox}
}

\newtcolorbox{resultbox}{
  breakable,
  sharp corners,
  colframe=black,
  colback=white,
  boxrule=0.4pt,
  left=2pt,
  right=2pt,
  top=2pt,
  bottom=2pt,
  fontupper=\linespread{0.7}\selectfont
}

\usepackage{changepage}

\title{
How Is LLM Reasoning Distracted by Irrelevant Context?\\An Analysis Using a Controlled Benchmark}

\author{
  Minglai Yang$^{1}$ \quad Ethan Huang$^{1}$ \quad Liang Zhang$^{1}$ \\
  \textbf{Mihai Surdeanu$^{1}$} \quad \textbf{William Wang$^{3}$} \quad
  \textbf{Liangming Pan$^{2,1}$}\thanks{\quad Corresponding author.} \\
  $^{1}$University of Arizona \quad $^{2}$MOE Key Lab of Computational Linguistics, Peking University \\
  $^{3}$University of California, Santa Barbara \\
  \texttt{\{mingly, ehuang68, liangzh, msurdeanu\}@arizona.edu} \\
  \texttt{liangmingpan@pku.edu.cn}\\
  \texttt{william@cs.ucsb.edu}
}

\begin{document}







\maketitle

\begin{abstract}

We introduce \textit{Grade School Math with Distracting Context} (GSM-DC\footnote{The code of our dataset and experiment can be viewed at \url{https://github.com/yminglai/GSM-DC}.}), a synthetic benchmark to evaluate Large Language Models' (LLMs) reasoning robustness against systematically controlled irrelevant context (IC). GSM-DC constructs symbolic reasoning graphs with precise distractor injections, enabling rigorous, reproducible evaluation. Our experiments demonstrate that LLMs are significantly sensitive to IC, affecting both reasoning path selection and arithmetic accuracy. Additionally, training models with strong distractors improves performance in both in-distribution and out-of-distribution scenarios. We further propose a stepwise tree search guided by a process reward model, which notably enhances robustness in out-of-distribution conditions. 

\end{abstract}

\section{Introduction}
Recent advances in Large Language Models (LLMs) have demonstrated reasoning capabilities across diverse tasks, notably in solving mathematical problems~\cite{cobbe2021gsm8k, lewkowycz2022solving, zhou2022least, yao2023tree}. Despite these advancements, LLMs are found to be less robust in reasoning~\cite{berglund2024reversalcursellmstrained,huang2024largelanguagemodelsselfcorrect, xu2024prideprejudicellmamplifies}. For example, the Flanker Task \cite{Eriksen1974EffectsON} in cognitive psychology shows that humans' responses become slower and less accurate with increased distractors. \citet{DBLP:conf/icml/ShiCMSDCSZ23} first revealed that LLMs similarly suffer performance degradation when irrelevant context is introduced, observing notable reductions in accuracy even with just a single distractor sentence added to math problems from the GSM8K dataset~\cite{cobbe2021gsm8k}. 



\begin{figure*}[ht]
    \centering
    \includegraphics[width=1.0\linewidth]{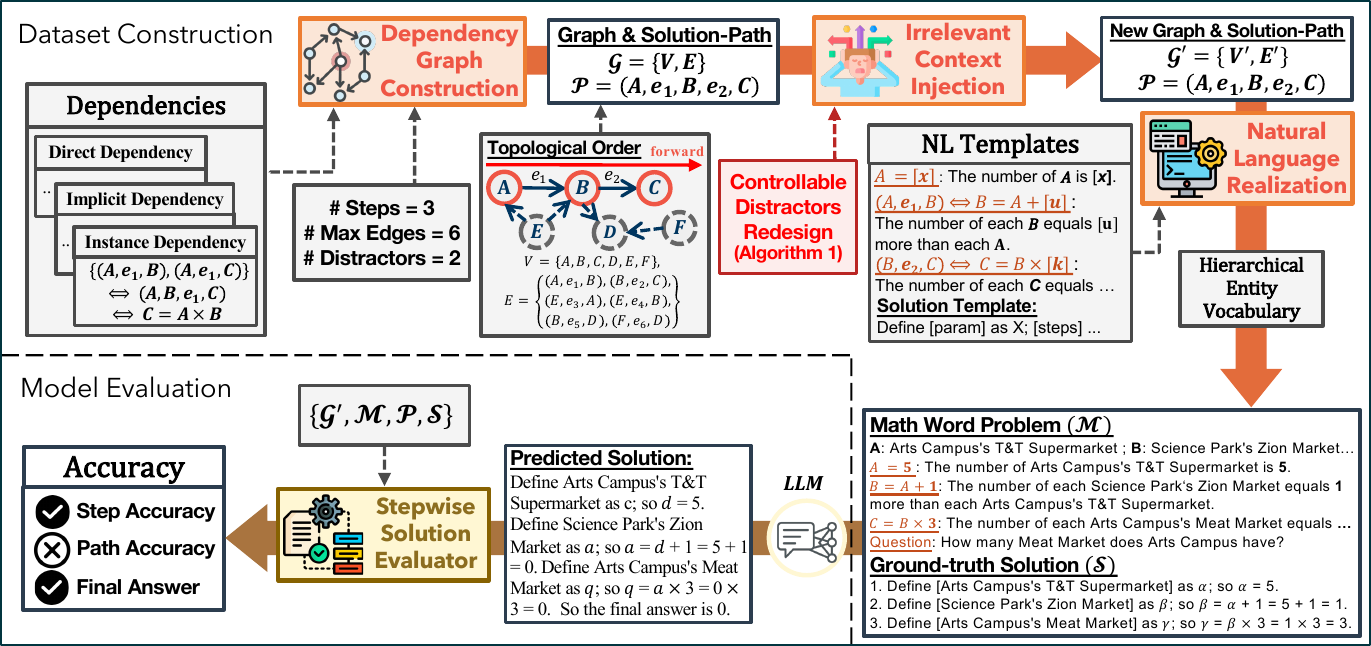}
    \caption{
    Overview of the \dataset{} framework: both generation and evaluation pipeline. The dataset construction process (orange) involves three key steps: (1) \textbf{\textcolor{myorange}{Dependency Graph Construction}} builds a symbolic DAG with a defined solution path via topological sort, (2) \textbf{\textcolor{myorange}{Irrelevant Context Injection}} adds controllable distractor nodes to increase reasoning complexity, and (3) \textbf{\textcolor{myorange}{Natural Language Realization}} converts the symbolic graph into a human-readable word problem and finds the solution following the solution path. The resulting instance is evaluated using a \textbf{\textcolor{myyellow}{Stepwise Solution Evaluator}} that computes Step Accuracy, Path Accuracy, and Extraction Answer Accuracy.}
    \label{fig:pipeline}
\end{figure*}

Prior work has not systematically explored the mechanisms underlying this sensitivity. \citet{DBLP:conf/icml/ShiCMSDCSZ23} employed only a single distractor, limited experiments to short reasoning chains, and omitted supervised fine-tuning and out-of-distribution (OOD) evaluations. Several important questions thus remain: How does varying the amount of IC affect robustness? Can robust reasoning be enhanced through supervised fine-tuning? How does the intensity of IC during training impact model performance in both in-distribution and OOD scenarios? Additionally, how can the above questions be qualitatively evaluated? 

To address these gaps, we introduce \dataset{}, a synthetic benchmark designed to enable precise control over both reasoning complexity and distractor structure. Problems in \dataset{} are represented as symbolic dependency graphs, where nodes correspond to intermediate quantities and edges represent symbolic operations. This structure facilitates: 1) the explicit injection of irrelevant context via off-path nodes and edges without affecting correct solutions; 2) adjustment of reasoning complexity by varying graph depth and structure; and 3) automatic evaluation of model outputs by aligning predictions with the correct reasoning path. 

Our dataset construction pipeline (\autoref{fig:pipeline}) involves generating symbolic dependency graphs, injecting distractors after determining the solution path, and transforming these into human-readable math word problems and solutions. We partition our dataset based on different problem complexities and distractor intensities, conduct various controlled experiments, and use automatic stepwise metrics measuring arithmetic correctness and distraction robustness. Our controlled experiments yield three main findings. First, model accuracy steadily decreases as distractor intensity rises. Second, continued pretraining substantially enhances reasoning robustness. Third, incorporating strong IC during training significantly boosts model resilience, showing superior performance across various distractor intensities in out-of-domain testing.  

To improve the model's robustness against IC, we propose a stepwise beam search algorithm guided by a Process Reward Model (PRM), which scores partial reasoning paths based on their alignment with valid solution trajectories. This approach further improves robustness by up to 6.29\% in out-of-domain conditions, highlighting reinforcement learning's potential to strengthen robustness against irrelevant context in model reasoning.

    \section{Related Work}
\vspace{-6pt}
\paragraph{Reasoning with Irrelevant Context}
\label{sec:reasoning_with_ic}
LLMs often struggle to reason accurately in the presence of irrelevant context (IC). Prior work has explored this vulnerability by introducing distractors into math problems. For example, GSM-IC \citep{DBLP:conf/icml/ShiCMSDCSZ23} appends irrelevant sentences to arithmetic questions but lacks control over distractor structure or complexity. GSMIR \citep{DBLP:journals/corr/abs-2408-10615} and MPN \citep{DBLP:conf/cikm/SongT24} use handcrafted prompting strategies to mitigate the effects of textual noise. \citet{anantheswaran2024cutting} generate adversarial math problems by adding irrelevant variables, showing significant performance drops and partial robustness gains through fine-tuning. However, their hand-crafted distractors risk introducing bias and lack structural control. Other studies~\citep{DBLP:journals/corr/abs-2404-03302,wang2025positionbiasmitigatesposition, xiao-etal-2025-mitigating,dong2025mmdocirbenchmarkingmultimodalretrieval} show that semantically similar but irrelevant documents can impair LLM performance, but RAG can improve robustness~\citep{jiang2024longragenhancingretrievalaugmentedgeneration,dong-etal-2024-mc,liu2024raginstructboostingllmsdiverse,wei2025surveylinkpredictionnary}. While these works expose LLMs’ sensitivity to IC, they provide limited control over distractor properties. In contrast, \dataset{} injects distractors into symbolic reasoning graphs, enabling stepwise evaluation. We further show that a reward-guided beam search improves robustness beyond standard fine-tuning.

\paragraph{Understanding LLM Reasoning}
LLM reasoning has received growing attention, leading to diverse efforts to improve performance on complex tasks. Recently, synthetic benchmarks such as GSM-$\infty$ \citep{DBLP:journals/corr/abs-2502-05252} and iGSM \citep{DBLP:journals/corr/abs-2407-20311} explored LLM reasoning under long-context and complex distractors. Unlike GSM-$\infty$ and iGSM,
our \textit{GSM-DC} explicitly controls irrelevant distractors within symbolic DAGs to systematically quantify the effects of irrelevant context. \citet{DBLP:journals/corr/abs-2404-05221} introduced AutoRace and the LLM Reasoners library to standardize reasoning evaluation. CoT prompting and in-context learning~\citep{xiong2025hsstarhierarchicalsamplingselftaught, yang2025utmathmathevaluationunit} have been shown to enhance logical reasoning \citep{DBLP:conf/emnlp/BertolazziGB24}, while other work highlights limitations in handling strict deductions \citep{DBLP:conf/llm4code/LiCXJLTLL24}. Recent methods such as ReAct \citep{DBLP:conf/iclr/YaoZYDSN023}, Tree-of-Thoughts \citep{DBLP:conf/nips/YaoYZS00N23}, and self-consistency decoding \citep{DBLP:conf/iclr/0002WSLCNCZ23} guide intermediate steps to improve solution quality. Beyond final-answer supervision, Process Reward Models (PRMs) \citep{DBLP:journals/corr/abs-2211-14275, DBLP:conf/iclr/LightmanKBEBLLS24, DBLP:journals/corr/abs-2412-06559, DBLP:journals/corr/abs-2409-12917, DBLP:journals/corr/abs-2402-06457} evaluate partial reasoning paths to promote more robust, interpretable, and aligned multi-step reasoning. Furthermore, Self-taught reasoners~\citep{zelikman2022starbootstrappingreasoningreasoning,huang2024largelanguagemodelsselfcorrect} reallocates sampling toward boundary problems via reward-guided difficulty estimation.
Finally, \citet{DBLP:conf/nips/ShiXWZZZTM23} showed that few-shot abduction boosts generalization with minimal supervision.

    \section{The \dataset{} Dataset}
\label{sec:graph_problem_gen}

To systematically investigate how LLMs reason under irrelevant context (IC), we require a framework that satisfies three desiderata: 1) fine-grained manipulation of IC, 2) precise control over reasoning difficulty, and 3) automatic evaluation of reasoning robustness.  Existing datasets (§\ref{sec:reasoning_with_ic}) like GSM-IC are manually built and rely on free-form outputs, lacking structural constraints and making stepwise evaluation impractical without manual checks.

We propose the \textit{Grade School Math with Distracting Context} (\textbf{\dataset{}}) benchmark—a controlled framework for systematically evaluating LLMs' reasoning under irrelevant context that meets the above criteria. 
Each math word problem in \dataset{} is represented as a directed acyclic graph (DAG), which allows us to 1) explicitly control irrelevant context by injecting distracting nodes and edges, 2) explicitly control reasoning difficulty by adjusting the graph size, and 3) automatically compute stepwise reasoning correctness by comparing model predictions to the ground-truth reasoning path. As illustrated in \autoref{fig:pipeline}, we construct the \dataset{} dataset in three steps:


\vspace{-0.2cm}
\paragraph{1) Dependency Graph Construction (§\ref{subsec: Dependency_Graph_Construction}):} To represent a math word problem, we build a symbolic dependency graph \(\mathcal{G}\) to capture the direct, implicit, and instance-level dependencies in the problem. We then identify a single correct reasoning path \(\mathcal{P}\) from the graph \(\mathcal{G}\) via topological sort.
\vspace{-0.65cm}
\paragraph{2) Irrelevant Context Injection (§\ref{subsec:Irrelevant_Context_Injection}):} We turn all nodes outside the reasoning path \(\mathcal{P}\) into distractors,
producing an augmented graph \(\mathcal{G}'\). 
This allows us to explicitly control the problem complexity (\textit{e.g.}, number of reasoning steps) and the intensity of irrelevant context (\textit{e.g.}, via the number and connectivity of distractor nodes).
\vspace{-0.2cm}
\paragraph{3) Natural Language Realization (§\ref{subsec:Natural_Language_Realization}):} We then convert the augmented graph \(\mathcal{G}'\) into a human-understandable math word problem \(\mathcal{M}\) by mapping each node to a real-world entity and rendering each edge into a statement. The ground-truth solution \(\mathcal{S}\) is then derived from the original reasoning path \(\mathcal{P}\).

\vspace{0.1cm}

As a result, each problem in the \dataset{} is represented as (\(\mathcal{G}'\), \(\mathcal{M}\), \(\mathcal{P}\), \(\mathcal{S}\)). This structured representation enables automatic stepwise evaluation (§\ref{subsec:Stepwise_Solution_Evaluator}) of LLMs' reasoning chain via the ground-truth path \(\mathcal{P}\). In the following, we will introduce the dataset construction pipeline in detail. 


\subsection{Dependency Graph Construction}
\label{subsec: Dependency_Graph_Construction}
Many grade-school math or logical reasoning problems involve quantities that are interrelated in various ways. These dependencies typically fall into three categories: 
1) \textit{Direct dependencies}, where one quantity is computed directly from another (\textit{e.g.}, if $R$ denotes the radius of a circle and $T$ its diameter, then $T = 2 \times R$);
2) \textit{Instance dependencies}, one entity is automatically reliant on another without explicitly stating that reliance. (\textit{e.g.}, ``Each shelf holds \(M\) books, and there are \(N\) shelves'') and 3) \textit{Implicit dependencies}, requiring aggregation or inference over multiple quantities (\textit{e.g.}, grouping cats and dogs as animals). 

To model these interrelations, we use the directed acyclic graph (DAG), denoted as \(\mathcal{G}\), where each \textit{node} denotes a quantity (\textit{e.g.}, Bob's pens) and each \textit{edge} represents the dependency between quantities (\textit{e.g.}, Alice has one more pen than Bob). We name \(\mathcal{G}\) as the \textit{dependency graph}. 
We use DAG because the acyclicity ensures that no quantity depends on itself, allowing a valid solution path \(\mathcal{P}\) to be recovered via topological sort. 

This structured graph-based representation forms the foundation for controlling reasoning complexity and enables injection of irrelevant context without affecting the original solution path \(\mathcal{P}\). 
Given inputs—reasoning steps \(rs\), maximum edges \(E\) and distractor count \(m\)—we generate a DAG by sampling nodes and edges, then extract the solution path \(\mathcal{P}\) of length \(rs\) via topological sort, and finally inject \(m\) controllable distractors (\S\ref{subsec:Irrelevant_Context_Injection}). All GSM-DC instances are guaranteed to be well-defined and solvable~\citep{tian2025vcsearchbridginggap} with a unique solution path \(\mathcal{P}\).

\subsection{Irrelevant Context Injection}
\label{subsec:Irrelevant_Context_Injection}

\begin{figure}[ht]
    \centering
    \includegraphics[width=1.0\linewidth]{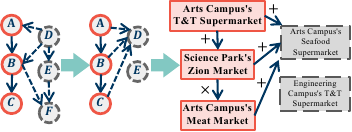}
    \caption{%
    Distractor construction in \dataset{}. After generating a DAG, we retain only the original topological path used in the solution (A → B → C). Distractor nodes are constructed by adding forward edges from solution nodes to unused parameters, preserving acyclicity. Since we control which unused parameters are included and their dependencies, D is the sum of A, B and C; E becomes an independent variable; F is excluded entirely.}
    \label{fig:distractor}
\end{figure}

To create a problem with irrelevant information, we augment the dependency graph by injecting distractor nodes while preserving the original solution path. As illustrated in \autoref{fig:distractor} and described in Algorithm~\ref{alg:distractor}, we start with a clean dependency graph \(\mathcal{G}\) and its solution path \(\mathcal{P}\). Unused nodes, which are not part of \(\mathcal{P}\), are selected and connected to existing nodes through forward-only edges, resulting in a new graph \(\mathcal{G}'\) that remains acyclic. 

Problem difficulty is primarily controlled by the number of reasoning steps \(rs\). To limit the problem complexity across instances, we constrain the input DAG \(\mathcal{G}\) to have at most \(E\) edges. Given such a fixed-scale graph and its solution path \(\mathcal{P}\), we inject \(m\) distractor nodes (none of which lie on \(\mathcal{P}\)) to produce the augmented graph \(\mathcal{G}'\) (Algorithm~\ref{alg:distractor}). 
Importantly, because the total graph scale is bounded by \(E\), longer reasoning steps occupy more of the graph structure, leaving fewer nodes and edges available for distractor injection. We vary \(m\in[m_{\min},m_{\max}]\) to define three distractor intensity levels (\textit{e.g.}, for \(rs=2\), light uses \(m\in[0,2]\), medium \(m\in[3,4]\), hard \(m\ge5\)). To ensure equal‐sized noise levels, we compute the empirical CDF of distractor levels \(z_i\) as \(\hat F_z(t)=\tfrac{1}{M}\sum_{i=1}^M\mathbb{I}(z_i\le t)\) and select \(m=\tau_k\) with \(\hat F_z(\tau_k)=\tfrac{k}{N}\) for \(k\sim\mathrm{Uniform}\{1,\dots,N\}\). You can see the full details are in Appendix~\ref{appendix:quan_irr_nodes}.

\begin{algorithm}[t]
\small
\caption{\textsc{InjectDistractors} (\autoref{fig:distractor})}
\label{alg:distractor}
\begin{algorithmic}[1]
\Require Directed acyclic graph $\mathcal{G}$, solution path $P$
\Ensure Augmented graph $\mathcal{G}'$ with $P$ preserved
\State $\mathcal{G}' \gets \mathcal{G}$ \Comment{work on a copy}
\State $\mathcal{R} \gets$ \textsc{UnusedParameters}$(\mathcal{G}', \mathcal{P})$
\While{$\mathcal{R} \neq \emptyset$}
    \State Sample batch $\mathcal{B}\subseteq\mathcal{R}$ with $|\mathcal{B}|=m$
    \ForAll{$\chi \in \mathcal{B}$}
        \State $\mathcal{R} \gets \mathcal{R}\setminus\{\chi\}$;\; $n \gets$ \textsc{NewNode}$(\chi)$
        \State \textsc{AddNode}$(\mathcal{G}', n)$ \Comment{$n$ is now a \emph{distractor}}
        \If{\textsc{IsUniqueTarget}$(\chi)$}
            \State \textsc{LabelIndependent}$(n)$ \Comment{$n$ has no parents}
            \State \textbf{continue}
        \EndIf
        \State Choose parent set $\mathcal{P}\in\{\mathcal{I}, \mathcal{C}\}$ with prob\ $q$
        \State \textsc{AddEdgesForward}$(\mathcal{G}', n, \mathcal{P}, \rho)$
        \State \textsc{LabelComputed}$(n)$
    \EndFor
\EndWhile
\State \Return $(\mathcal{G}', P)$
\end{algorithmic}
\end{algorithm}

\subsection{Natural Language Realization}
\label{subsec:Natural_Language_Realization}

Once the dependency graph \(\mathcal{G}\) is constructed and augmented as \(\mathcal{G}'\), we instantiate it into natural language. Each node is mapped to an entity (\textit{e.g.}, ``Arts Campus's T\&T Supermarket'') from the hierarchical entity vocabulary of the GSM8K dataset, and each edge is rendered using a templated relational statement (\textit{e.g.}, ``the number of Zion Markets is 1 more than the number of T\&T Supermarkets'')\footnote{We adopt the hierarchical entity vocabulary and templated relational statements introduced in \cite{DBLP:journals/corr/abs-2407-20311}.}. These templates capture the underlying dependencies while maintaining simple, readable language. 

To form the math problem \(\mathcal{M}\), we concatenate natural-language realizations of edges along the solution path, ending with a question about the final node. Distractors are rendered as unrelated sentences and shuffled with relevant content.

\begin{figure}[ht]
  \centering
  \includegraphics[width=1.0\linewidth]{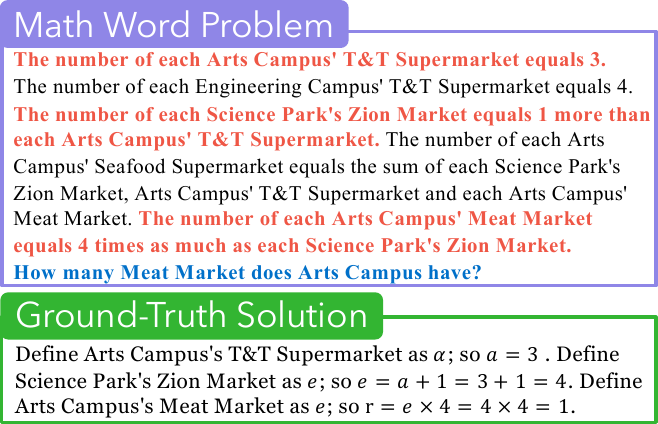}
    \caption{The final reasoning problem constructed from the example in \autoref{fig:distractor}. \textcolor{mwpred}{\textbf{Red}} sentences are the relevant facts on the gold reasoning path $P$; \textbf{black} sentences are the injected IC; the \textcolor{mwpblue}{\textbf{blue}} sentence is the final question.
}
  \label{fig:sample}
\end{figure}

Alongside the natural language (NL) problem \(\mathcal{M}\), we generate its corresponding NL solution \(\mathcal{S}\) based on the ground-truth reasoning path \(\mathcal{P}\). The solution \(\mathcal{S}\) sequentially defines variables for each node along the path \(\mathcal{P}\) and applies the dependencies. An example of the NL problem is given in \autoref{fig:sample}. 

The generated natural language solution provides a templated Chain-of-Thought (CoT) reasoning trace, which can be parsed to automatically evaluate the stepwise reasoning correctness.

While the underlying DAGs in \dataset{} are randomly generated, the resulting math word problems are not. We use manually constructed semantic hierarchies to ensure each problem is semantically coherent. This allows us to focus on isolating and evaluating reasoning robustness under controlled distractor settings. For further discussion on how this synthetic design compares to human-authored problems and why it is appropriate for our controlled experiments, see Appendix~\ref{appendix:synthetic}\footnote{We consider arithmetics mod 5 to avoid errors from computation involving large numbers. LLMs can handle arithmetic via external tools~\citep{schick2023toolformerlanguagemodelsteach, paranjape2023artautomaticmultistepreasoning}.}.

\subsection{Stepwise Solution Evaluator}
\label{subsec:Stepwise_Solution_Evaluator}
After constructing \dataset{}, we build a stepwise solution evaluator to automatically evaluate LLM-generated solutions. For each problem and predicted solution, we report three \emph{binary} scores; for each, a value of 1 is awarded only when the stated criterion is fully satisfied. 
\vspace{-6pt}
\paragraph{$\bullet$ Step Accuracy (SAcc):}  
          Our symbolic parser reads the model’s chain‑of‑thought and executes every intermediate equation in topological order.  
        SAcc = 1 iff \emph{all} equations are arithmetically correct \emph{and} each step references only symbols that have already been defined. We enforce node-level alignment in the parser (not strict sequence matching). 
          This strict all‑or‑nothing formulation avoids inflating performance with partially correct derivations.
\vspace{-19pt}
\paragraph{$\bullet$ Path Accuracy (PAcc):}  
          To quantify \emph{distraction robustness} we check whether the model confines its reasoning to the augmented dependency graph \(\mathcal{G}'\) after injecting irrelevant context.  
        PAcc = 1 iff (i) every required dependency on \(P\) is present, and (ii) no irrelevant node is used in the computation of any step on \(P\). Similar to SAcc, evaluation is performed via \emph{node-level alignment}, not strict sequence matching: valid steps may appear in any order, and redundant or extra steps are permitted, so long as they do not interfere with the correct solution path \(P\). PAcc is a relaxation of SAcc as it only requires stepwise reasoning to be correct, but not the associated values themselves.
\vspace{-6pt}
\paragraph{$\bullet$ Extraction Answer Accuracy (EAcc):}  
            To capture final‐answer correctness, EAcc = 1 iff the model’s extracted answer exactly matches the ground truth. We report EAcc only for prompting, but our focus remains on SAcc and PAcc.
            
\vspace{0.1cm}
We evaluate these metrics over a large set of problems and report each as the percentage (\%) of instances achieving a score of 1.

    \section{Experiments}
\label{sec:experiments}


\subsection{Impact of Irrelevant Context}
\label{subsec:ic_impact}
To systematically analyze how irrelevant context (IC) affects LLM reasoning, we conduct controlled experiments by injecting varying numbers of irrelevant context (\(m = 1\text{–}15\)) into math word problems \(\mathcal{M}\) drawn from \dataset{} (§\ref{sec:graph_problem_gen}). 
We evaluate performance across four levels of reasoning steps, denoted \( rs \in \{2, 3, 4, 5\} \), and sample 100 instances per condition to ensure statistical stability.

We benchmark six instruct models: Grok-3-Beta, GPT-4.1, GPT-4o-mini, LLaMA-3.3-70B, LLaMA-3.1-8B and LLaMA-3.2-1B. We employ a five-shot prompting strategy enhanced with a structured \textit{Background} section (Appendix~\ref{appendix:prompt}) that explicitly encodes relevant dependencies to guide reasoning. 
Model performance is assessed using three metrics using \textit{Stepwise Solution Evaluator}, SAcc, PAcc and EAcc, which together capture reasoning correctness, robustness to distractors, and output correctness (§\ref{subsec:Stepwise_Solution_Evaluator}). This decomposition allows us to isolate the specific ways in which irrelevant context degrades model performance.


\begin{figure*}[h!]
    \centering
    \begin{subfigure}[t]{0.49\textwidth}
        \centering  \includegraphics[width=\textwidth]{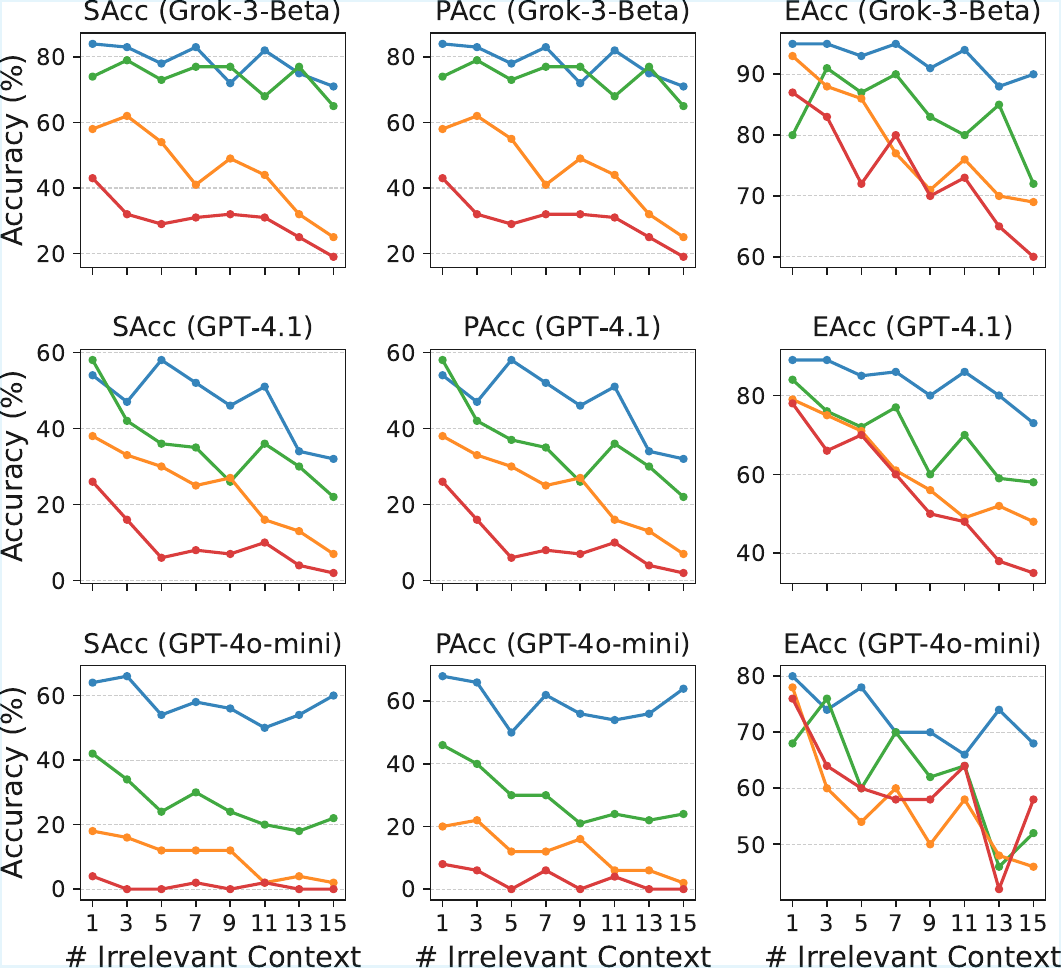}
        \label{fig:degrade1}
    \end{subfigure}
    \hfill
    \begin{subfigure}[t]{0.492\textwidth}
        \centering
\includegraphics[width=\textwidth]{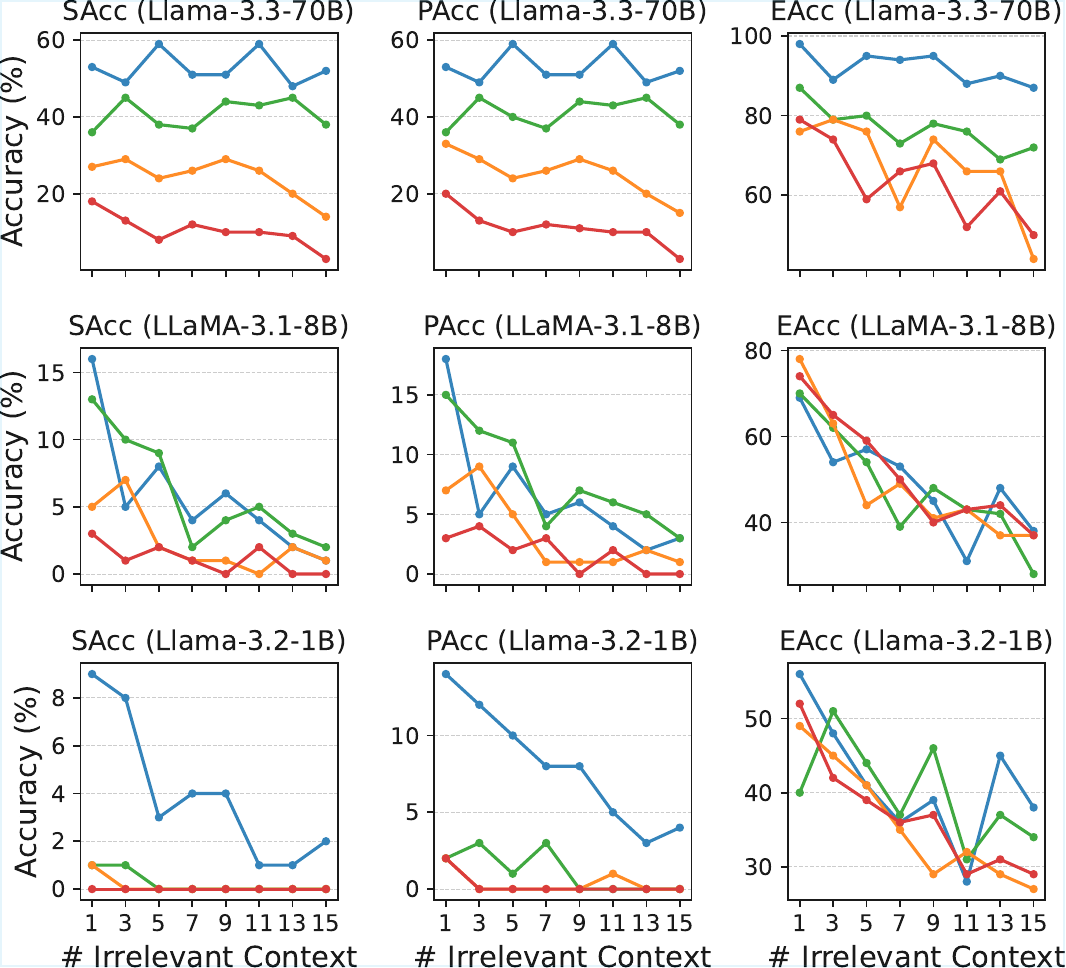}
        \label{fig:degrade2}
    \end{subfigure}
\caption{
Step-wise accuracy under increasing irrelevant context (IC) for four models, evaluated across reasoning steps $rs \in \{2,3,4,5\}$. 
We adopt a 5-shot prompting strategy with background information. 
Each curve corresponds to a specific reasoning step: \rslegend.
}
\vspace{-0.3cm}
\label{fig:degrade}
\end{figure*}

\researchfinding{\textbf{I:} LLMs' reasoning performance degrades with increasing irrelevant context.}

As shown in \autoref{fig:degrade}, all six models exhibit a clear degradation in reasoning accuracy as the number of irrelevant context increases. For instance, at a fixed reasoning depth of \(rs{=}5\), Grok-3-Beta’s step accuracy drops from 43\% with one irrelevant context to just 19\% under fifteen irrelevant context. GPT‑4.1 exhibits an even steeper decline at the same depth, falling from 26\% to 2\%.

All three evaluation metrics—step accuracy (SAcc), path accuracy (PAcc), and extraction accuracy (EAcc)—exhibit similar downward trends as irrelevant context increases. Extraction accuracy (EAcc) remains relatively high, because our solution parser enforces a strict Chain-of-Thought format (§\ref{subsec:Stepwise_Solution_Evaluator}) that models learn to follow through five-shot prompting. As a result, EAcc is less sensitive to distraction compared to SAcc and PAcc, which more directly assess reasoning fidelity and resistance to irrelevant information. 



\researchfinding{\textbf{II:} Irrelevant context degrades accuracy more steeply at greater reasoning depths.}

To analyze how irrelevant context (IC) interacts with reasoning complexity, we study the error rate \(E(m; rs)\) as a function of distractor count \(m\) and reasoning depth \(rs\). We find it roughly follows a power-law trend: \(E(m; rs) \propto m^{\delta(rs)}\), where \(\delta(rs)\) reflects a model’s IC sensitivity. As shown in \autoref{fig:degrade}, error increases with \(m\), and the degradation steepens with deeper reasoning.

For instance, Grok-3-Beta’s exponent grows from \(\delta \approx 0.11\) at \(rs{=}2\) to \(\delta \approx 0.49\) at \(rs{=}5\), indicating greater vulnerability at deeper depths. GPT‑4.1 shows a similar slope but higher baseline error, suggesting that reasoning depth governs \(\delta(rs)\), while model capacity sets the vertical intercept—\textit{i.e.}, robustness under minimal distraction. These findings highlight the need to jointly consider reasoning complexity and IC sensitivity when designing robust LLMs. 

\subsection{Training with Different Strategies}
\label{subsec:finetune_strategy}

The results so far focus on inference-time behavior: models are prompted to reason through irrelevant context (IC) without being explicitly trained on it. However, since we do not have access to the original training data of these models, it is unclear whether their observed robustness (or lack thereof) stems from genuine generalization or incidental exposure to similar patterns during pretraining. To disentangle this, we perform controlled experiments that explicitly expose models to varying degrees of IC and reasoning complexity.

First, we conduct controlled experiments on \dataset{} with varying reasoning steps. We first mimic the distribution in \textsc{GSM-IC} by training on examples with 2–7 reasoning steps, then evaluate on harder problems with up to 22 steps. As shown in Appendix~\ref{appendix:gsm_ic}, performance drops sharply once the test depth exceeds the training horizon, suggesting that models fail to generalize if they trained with shallow reasoning samples.

To address this, we expand the training set to include examples up to \(rs{=}15\), ensuring exposure to both long reasoning chains and varying levels of irrelevant context. All finetuned models in this section are trained on this broader distribution and evaluated on both \textit{in-distribution} (\(rs \le 15\)) and \textit{out-of-distribution} (\(rs > 15\)) samples.




\begin{figure}[h!]
    \centering
    \includegraphics[width=0.46\textwidth]{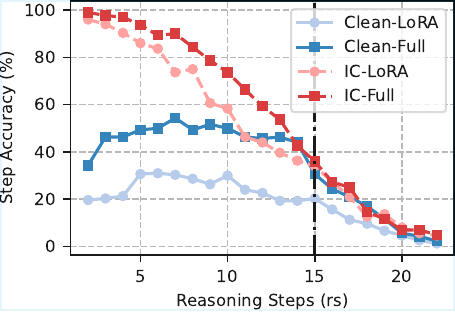}
    \caption{Step accuracy of models trained with Clean or IC data using LoRA or continued pretraining.}
    \label{fig:full_lora_clean}
\end{figure}

\researchfinding{\textbf{III:} Continued pretraining enhances robustness even without access to IC samples.}
Building on this controlled training setup, we investigate how different finetuning strategies affect reasoning robustness under irrelevant context.
Specifically, we compare continued pretraining (full finetuning) and LoRA finetuning for reasoning robustness using a 30K-sample training set, which we select based on empirical scaling trends analyzed in Appendix~\ref{appendix:data_scaling}. As shown in \autoref{fig:full_lora_clean}, continued pretraining confers strong robustness even without IC supervision, substantially outperforming LoRA on clean data. With IC training, the gap narrows, but continued pretraining remains consistently more robust across reasoning depths. Based on this, we fixed continued pretraining 30K-samples for all subsequent experiments.

\begin{table}
\centering
\centering
\setlength{\tabcolsep}{3pt}
\footnotesize
\resizebox{\columnwidth}{!}{%
\begin{tabular}{r|ccc|ccc|ccc}
\toprule
\multirow{2}{*}{\textbf{$rs$}} 
& \multicolumn{3}{c|}{\textbf{Clean}} 
& \multicolumn{3}{c|}{\textbf{Clean+IC}} 
& \multicolumn{3}{c}{\textbf{IC}} \\
\cmidrule(lr){2-4}\cmidrule(lr){5-7}\cmidrule(lr){8-10}
& SAcc & PAcc & EAcc & SAcc & PAcc & EAcc & SAcc & PAcc & EAcc \\
\midrule
$\leq 15$ & 35.9 & 41.3 & 68.7 & 70.0 & 71.2 & 84.9 & \textbf{73.2} & \textbf{74.7} & \textbf{87.3} \\
16        & 22.0 & 22.7 & 57.7 & 32.0 & 32.0 & 66.0 & \textbf{33.3} & \textbf{33.3} & \textbf{66.3} \\
17        & 21.0 & 21.0 & 55.3 & 23.0 & 23.0 & 61.7 & \textbf{20.7} & \textbf{21.3} & \textbf{66.7} \\
18        & 13.0 & 13.0 & 56.0 & 15.7 & 15.7 & 48.7 & \textbf{16.7} & \textbf{16.7} & \textbf{53.3} \\
19        & 13.7 & 13.7 & 57.0 & 13.3 & 13.3 & 49.3 & \textbf{15.0} & \textbf{15.0} & \textbf{60.3} \\
20        & 9.0  & 9.0  & 48.3 & 8.3  & 8.3  & 52.3 & \textbf{10.0} & \textbf{10.0} & \textbf{52.7} \\
21        & 7.7  & 7.7  & 47.7 & 8.7  & 8.7  & 48.0 & \textbf{8.7}  & \textbf{8.7}  & \textbf{49.3} \\
22        & 6.0  & 6.0  & 45.0 & 5.3  & 5.3  & 46.7 & \textbf{6.3}  & \textbf{6.3}  & \textbf{47.3} \\
\bottomrule
\end{tabular}%
}
\caption{SAcc, PAcc, and EAcc under different training regimes: Clean, Clean+IC, and IC—evaluated on the full GSM-DC dataset \emph{with IC}. EAcc can be misleadingly high when models guess the answer without following the correct path, motivating our focus on SAcc/PAcc.}

\label{tab:basic_sft}
\end{table}

\subsection{Control of Training Data}
\label{subsec:train_with_ic}

\researchfinding{\textbf{IV:} Training with irrelevant context improves robustness most effectively.}

As shown in \autoref{tab:basic_sft}, the model trained on IC consistently achieves the highest SAcc and PAcc across all \textit{rs}. The model trained on Clean+IC data performs slightly worse, while the Clean model lags behind both. These results suggest that training solely on IC leads to stronger robustness, because of increased exposure to IC during learning.

\begin{figure}[htbp]
    \centering
    \includegraphics[width=0.46\textwidth]{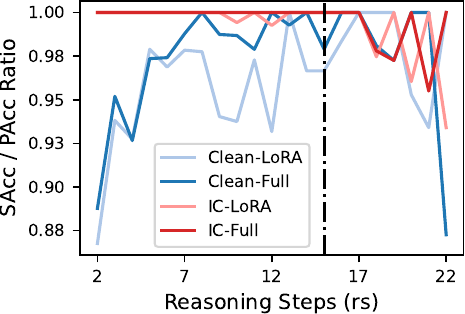}
    \caption{Step accuracy (\%) for models trained with \textit{Clean} or \textit{IC} using LoRA or continued pretraining. 
    }
    \label{img:delta}
\end{figure}

The clean model performs worse on questions with IC, even under in-distribution (ID) settings. To better understand this limitation, we examine the gap $\Delta(\textsc{SAcc}, \textsc{PAcc})$, represented as the ratio between SAcc and PAcc (\autoref{img:delta}). A lower ratio indicates a larger gap—arithmetic errors occurring even when the reasoning path is correct. The model trained on Clean data consistently shows a higher $\Delta$, suggesting that IC affects not only reasoning path selection, but also arithmetic execution. These findings reveal that IC broadly disrupts reasoning, and that training with IC-injected examples leads to more robust models.

\researchfinding{\textbf{V:} Training with challenging irrelevant context leads to the strongest robustness and generalization across all pretraining settings.}
Having established that exposure to irrelevant context during training improves robustness, we now investigate whether the intensity of such context further influences generalization. In particular, we test whether training on harder, more distracting IC leads to greater robustness on out-of-distribution (OOD) reasoning problems.
Based on the method described in \S\ref{sec:graph_problem_gen}, we construct two main versions of the \dataset{} benchmark for evaluation:
\vspace{-0.2cm}
\paragraph{\dataset{}-Clean:} For each reasoning step \( rs \in [2, 22] \cap \mathbb{Z} \), we sample 300 clean dependency graphs without injecting any IC. Each graph contains a unique solution path \(\mathcal{P}\) and no distractor nodes. This clean subset comprises 6{,}300 math problems.
\vspace{-0.2cm}
\paragraph{\dataset{}-with-IC:} To study robustness under IC, we generate IC variants of the clean graphs by injecting distractors following the procedure (§\ref{subsec:Irrelevant_Context_Injection}). For each reasoning step, we sample 100 graphs under each of three IC intensity levels: \textsc{Light-IC}, \textsc{Medium-IC}, and \textsc{Hard-IC}, while keeping the reasoning path \(\mathcal{P}\) fixed. Each subset thus contains 2{,}100 problems (100 per step), resulting in a total of 6{,}300 problems across all IC levels. 

To evaluate how IC difficulty affects training, we compare five regimes: \textsc{Clean}, \textsc{Light / Medium / Hard-IC}, and \textsc{Mix-IC}. As shown in \autoref{tab:control_noise_sft} and \autoref{tab:finegrain}, \textsc{Hard-IC} yields the best SAcc across all in-distribution and OOD settings, regardless of IC presence or difficulty.

\begin{table}[ht]
\centering
\centering
\setlength{\tabcolsep}{4pt}
\footnotesize
\resizebox{\columnwidth}{!}{%
\begin{tabular}{lccc|ccc}
\toprule
\multirow{2}{*}{\makecell{Training\\IC Level}} 
  & \multicolumn{3}{c|}{\textbf{Testing w/ IC (SAcc)}} 
  & \multicolumn{3}{c}{\textbf{Testing w/o IC (SAcc)}} \\
\cmidrule(lr){2-4}\cmidrule(lr){5-7}
 & ID & OOD & All & ID & OOD & All \\
\midrule
\textsc{Clean}   & 35.91 & 13.19 & 32.36 & 81.95 & 17.05 & 60.32 \\
\textsc{Light-IC} & 64.79 &  6.90 & 46.57 & 67.33 &  7.09 & 46.56 \\
\textsc{Medium-IC} & 65.79 & 7.23 & 47.44 & 69.39 & 9.95 & 50.38 \\
\textsc{Hard-IC}   & \textbf{77.95} & \textbf{18.57} & \textbf{59.48} & \textbf{82.30} & \textbf{19.86} & \textbf{61.21} \\
\textsc{Mix-IC}  & 73.23 & 15.33 & 57.86 & 78.09 & 15.62 & 57.38 \\
\bottomrule
\end{tabular}%
}
\caption{Step Accuracy (\%) under different training IC difficulties, evaluated across test IC conditions.}
\label{tab:control_noise_sft}
\end{table}

\begin{table}
\centering
\centering
\setlength{\tabcolsep}{4pt}
\footnotesize
\resizebox{\columnwidth}{!}{%
\begin{tabular}{lccc|ccc}
\toprule
\multirow{2}{*}{\makecell{Training\\IC Level}} & \multicolumn{3}{c|}{\textbf{ID Test SAcc}} & \multicolumn{3}{c}{\textbf{OOD Test SAcc}} \\
\cmidrule(lr){2-4}\cmidrule(lr){5-7}
 & Light & Medium & Hard & Light & Medium & Hard \\
\midrule
\textsc{Light-IC} & 67.21 & 66.57 & 60.57 & 8.14 & 7.29 & 5.28 \\
\textsc{Medium-IC} & 68.14 & 66.07 & 63.14 & 8.71 & 8.43 & 4.57 \\
\textsc{Hard-IC} & \textbf{78.36} & \textbf{79.21} & \textbf{76.28} & \textbf{22.7} & \textbf{18.43} & \textbf{14.57} \\
\textsc{Mix-IC} & 74.71 & 75.07 & 69.93 & 17.7 & 16.57 & 11.28 \\
\bottomrule
\end{tabular}%
}
\caption{Step Accuracy (\%) per test IC difficulty. All models are trained with a specific IC difficulty.}
\label{tab:finegrain}
\end{table}

These findings indicate that exposure to adding challenging distractors (\textsc{Hard-IC}) is the most effective training strategy for enhancing model robustness and generalization performance. Intriguingly, \textsc{Mix-IC}, despite incorporating distractor diversity, consistently underperformed \textsc{Hard-IC}, suggesting that distractor difficulty, rather than variety, is the primary driver of improvement. The advantage of \textsc{Hard-IC} over \textsc{Clean}, particularly under test-time IC conditions, further reinforces the utility of IC augmentation, specifically with high-difficulty examples, for fostering robust reasoning.

    \section{Improving Model Robustness Against Irrelevant Context}
\label{sec:tree_search}

\begin{figure}[htbp]
    \centering
    \includegraphics[width=1.01\linewidth]{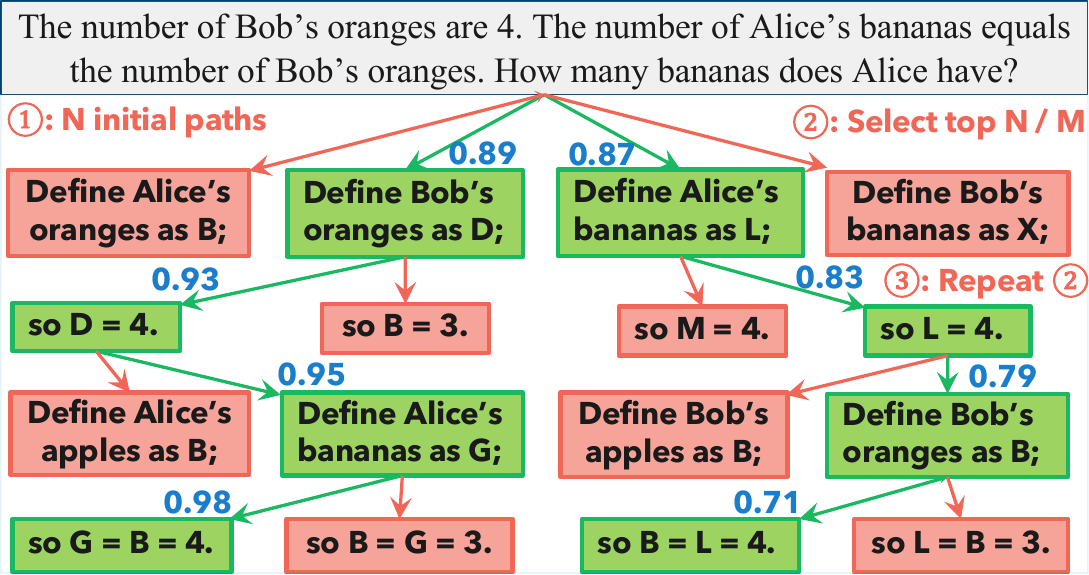}
    \caption{An overview of the ToT algorithm with $N = 4$ and $M = 2$. Green nodes indicate those that were scored highly by the PRM and thus expanded in subsequent iterations, while red nodes were not selected as candidate nodes for the next step. After the algorithm terminates, the leftmost node is scored the highest and thus that reasoning path is chosen as the final answer.} 
    \label{fig:treesearch}
\end{figure}

The previous section (\S\ref{sec:experiments}) demonstrate that LLMs are highly sensitive to irrelevant context (IC), and that continued pretraining with challenging IC-injected examples alone can substantially improve robustness. However, even with the strongest continued pretraining configurations (\textit{e.g.}, \textsc{Hard-IC}), model performance still degrades significantly on out-of-distribution (OOD) reasoning steps. This raises the question of how robustness can be further improved at test time.





\researchfinding{\textbf{VI:} PRM-guided tree search preserves ID accuracy while consistently boosting OOD robustness across all IC training levels.}

\begin{table}
\centering
\centering
\setlength{\tabcolsep}{4pt}
\footnotesize
\resizebox{\columnwidth}{!}{%
\begin{tabular}{lccc|ccc}
\toprule
\multirow{3}{*}{\makecell{Training\\IC Level}} 
  & \multicolumn{3}{c|}{\textbf{ID SAcc}} 
  & \multicolumn{3}{c}{\textbf{OOD SAcc}} \\
\cmidrule(lr){2-4} \cmidrule(lr){5-7}
  & \makecell{w/o} & \makecell{w/} & $\Delta$ 
  & \makecell{w/o} & \makecell{w/} & $\Delta$ \\
  & \makecell{PRM} & \makecell{PRM} & 
  & \makecell{PRM} & \makecell{PRM} & \\
\midrule
\textsc{Light-IC}   & 64.79 & \textbf{66.10} & \cellcolor{green!15}{\textcolor{green!60!black}{\textbf{+1.31}}} 
                   & 6.90  & \textbf{9.59}  & \cellcolor{green!15}{\textcolor{green!60!black}{\textbf{+2.69}}} \\
\textsc{Medium-IC}  & 65.79 & \textbf{70.05} & \cellcolor{green!15}{\textcolor{green!60!black}{\textbf{+4.26}}} 
                   & 7.23  & \textbf{13.52} & \cellcolor{green!15}{\textcolor{green!60!black}{\textbf{+6.29}}} \\
\textsc{Hard-IC}    & 77.95 & \textbf{79.48} & \cellcolor{green!15}{\textcolor{green!60!black}{\textbf{+1.53}}} 
                   & 18.57 & \textbf{24.17} & \cellcolor{green!15}{\textcolor{green!60!black}{\textbf{+5.60}}} \\
\textsc{Mix-IC}     & 73.23 & \textbf{75.81} & \cellcolor{green!15}{\textcolor{green!60!black}{\textbf{+2.58}}} 
                   & 15.33 & \textbf{19.06} & \cellcolor{green!15}{\textcolor{green!60!black}{\textbf{+3.73}}} \\
\textsc{Clean}      & 35.91 & \textbf{36.38} & \cellcolor{green!15}{\textcolor{green!60!black}{\textbf{+0.47}}} 
                   & 13.19 & \textbf{15.76} & \cellcolor{green!15}{\textcolor{green!60!black}{\textbf{+2.57}}} \\
\bottomrule
\end{tabular}%
}
\caption{The Step Accuracies of the models trained with different IC levels without and with PRM.}
\label{tab:prm_main}
\end{table}

Our Tree of Thoughts (ToT) algorithm addresses complex reasoning problems by combining tree search with the step-by-step inference capabilities of large language models (LLMs). As illustrated in \autoref{fig:treesearch}, ToT not only uses an LLM to propose candidate reasoning steps, but also integrates a Process Reward Model (PRM) to evaluate and guide the search process. Given a partial reasoning path \( h_{1:t} \), the PRM assigns a reward \( R(h_{1:t}) \) indicating the quality of reasoning up to step \( t \). Leveraging a synthetic dataset, we systematically inject irrelevant context (IC) and arithmetic errors into selected reasoning paths. These negative examples are used to train the PRM to distinguish valid reasoning trajectories from those corrupted by irrelevant context (IC) and wrong arithmetic calculations enabling the model to prioritize more accurate and robust solutions during search.


Through our experiments, we found that the measured accuracy, both SAcc and PAcc, for the in-distribution case with and without a PRM were similar. Furthermore, in the OOD case, the accuracy we measured was significantly improved when a PRM was used. The results suggest that using a PRM preserves model performance in ID tasks, while also allowing the model to generalize its responses to OOD tasks. As can be seen from \autoref{tab:prm_main}, the model trained with hard IC performs the greatest, and supplementing it with a PRM significantly improves its accuracy.

    \section{Conclusion}
We present \textbf{\dataset{}}, a controlled benchmark for rigorous evaluation and improving the robustness of LLM reasoning in the presence of systematically injected irrelevant context (IC). By framing math problems as symbolic DAGs, \dataset{} enables precise control over reasoning complexity and distractor structure, along with automatic stepwise evaluator. Our experiments reveal that: 1) LLM accuracy degrades as distractor count increases, with the error roughly following a power-law trend whose exponent grows with reasoning depth; 
2) IC affects not only reasoning path selection, but also arithmetic execution; 
3) Training with challenging IC, combined with continued pretraining, yields the strongest robustness across both in-distribution and out-of-distribution settings, consistently outperforming LoRA finetuning under clean and noisy conditions. Finally, we show that reasoning robustness can be further improved at inference time using beam search with PRM, which boosts OOD step accuracy by up to 6.29\%. Together, these findings position \dataset{} as both a diagnostic tool for analyzing IC sensitivity and a foundation for developing robust training and inference time strategies for language models reasoning.


\section*{Limitations}
\label{sec:limitations}

\dataset{} provides a controlled environment for probing LLM reasoning, combining symbolic DAGs with natural-language templates inspired by datasets like iGSM~\cite{DBLP:journals/corr/abs-2407-20311}. To enhance linguistic diversity and realism, we designed a hierarchical vocabulary system derived from GSM8K~\cite{cobbe2021gsm8k} and constructed templated prompts with varied surface forms. While this approach balances control and naturalness, the use of templates still limits full linguistic expressiveness. To address this, we plan to expand the benchmark with more diverse natural-language realizations sampled from real corpora and support more flexible arithmetic reasoning. The current reasoning depth is capped at 22 operations; we are generating new tiers with 30+ steps to explore long-horizon compositionality. While we benchmark six models—Grok-3-Beta, GPT-4.1, GPT-4o-mini, LLaMA-3.3-70B, LLaMA-3.1-8B, and LLaMA-3.2-1B—all training experiments are conducted solely on LLaMA-3.2-1B using a 30K-sample dataset (see Appendix~\ref{appendix:gpu}) due to computational constraints. Finally, we aim to include faithfulness and bias diagnostics—such as explanation consistency and demographic sensitivity—to ensure that robustness gains translate into safe and trustworthy reasoning.
    \section*{Acknowledgement}

This research uses only synthetic data and does not involve human subjects or sensitive information. All models and experiments comply with the licenses of publicly available tools. We support responsible AI research and have prioritized transparency and reproducibility throughout this work.

    \bibliography{custom}

    \clearpage

\appendix
\section*{Content of Appendix}
\begin{itemize}[noitemsep, topsep=0pt]
    \item[\ref{appendix:dataset}] Dataset Samples
    \begin{itemize}
        \item[\ref{appendix:finetune}] Training Dataset with Different Level of IC for Finetuned Model
        \item[\ref{appendix:prompt}] Prompts for Closed-Sourced Models
    \end{itemize}
    \item[\ref{appendix:quan_irr_nodes}] Empirical Noise Stratification
    \item[\ref{appendix:synthetic}] Clarifying the Role of Synthetic Problem Generation
    \item [\ref{appendix:gsm_ic}] Operation‑Range Bias and Our Train/Test Protocol
    \item [\ref{appendix:prm}] Process Reward Models
    \item [\ref{appendix:additional}] Additional Analyses
    \item [\ref{appendix:gpu}] Finetuning Details
    \item [\ref{appendix:future}] Future Work
\end{itemize}

\section{Dataset Samples}
\label{appendix:dataset}

\subsection{Training Dataset with Different IC for Finetuned Model}
For models that have been finetuned on mathematical reasoning tasks, we provide the question directly, omitting any system or instruction prompt.
\label{appendix:finetune}
\begin{tcolorbox}[
    breakable,
    colframe=black!75,
    colback=gray!10,
    coltitle=white,
    colbacktitle=black!80,
    title=\textbf{Light-IC Sample (Operations = 2)},
    fontupper=\small,
    fonttitle=\bfseries
]
$\blacktriangleright$ \textbf{Input:}\\
\texttt{\textbf{The number of each Arts Campus's T\&T Supermarket equals 3. The number of each Science Park's Zion Market equals 1 more than each Arts Campus's T\&T Supermarket.} The number of each Engineering Campus's Zion Market equals each Engineering Campus's T\&T Supermarket. \textbf{How many Zion Market does Science Park have?}}\\

$\blacktriangleright$ \textbf{Output:}\\
Define Arts Campus's T\&T Supermarket as e; so e = 3. Define Science Park's Zion Market as w; so w = e + 1 = 3 + 1 = 4.
\end{tcolorbox}

\begin{tcolorbox}[
    breakable,
    colframe=black!75,
    colback=gray!10,
    coltitle=white,
    colbacktitle=black!80,
    title=\textbf{Medium-IC Sample (Operations = 2)},
    fontupper=\small,
    fonttitle=\bfseries
]
$\blacktriangleright$ \textbf{Input:}\\
\texttt{\textbf{The number of each Arts Campus's T\&T Supermarket equals 3.} The number of each Arts Campus's La Michoacana Meat Market equals 4. 
The number of each Preparatory School District's La Michoacana Meat Market equals 3 more than the difference of each Science Park's T\&T Supermarket and each Science Park's La Michoacana Meat Market. \textbf{The number of each Science Park's Zion Market equals 1 more than each Arts Campus's T\&T Supermarket.} The number of each Engineering Campus's Zion Market equals each Engineering Campus's T\&T Supermarket. \textbf{How many Zion Market does Science Park have?}}\\

$\blacktriangleright$ \textbf{Output:}\\
Define Arts Campus's T\&T Supermarket as e; so e = 3. Define Science Park's Zion Market as w; so w = e + 1 = 3 + 1 = 4.
\end{tcolorbox}

\begin{tcolorbox}[
    breakable,
    colframe=black!75,
    colback=gray!10,
    coltitle=white,
    colbacktitle=black!80,
    title=\textbf{Hard-IC Sample (Operations = 2)},
    fontupper=\small,
    fonttitle=\bfseries
]
$\blacktriangleright$ \textbf{Input:}\\
\texttt{The number of each Arts Campus's La Michoacana Meat Market equals 4. \textbf{The number of each Arts Campus's T\&T Supermarket equals 3.} The number of each Arts Campus's Seafood City Supermarket equals 2 more than each Science Park's Zion Market. The number of each Preparatory School District's Zion Market equals each Engineering Campus's Seafood City Supermarket. The number of each Science Park's Seafood City Supermarket equals the sum of each Science Park's La Michoacana Meat Market and each Science Park's T\&T Supermarket. The number of each Preparatory School District's Seafood City Supermarket equals 4 more than the sum of each Science Park's Zion Market, each Arts Campus's T\&T Supermarket and each Arts Campus's Seafood City Supermarket. The number of each Arts Campus's Zion Market equals the sum of each Science Park's T\&T Supermarket, each Arts Campus's T\&T Supermarket and each Engineering Campus's La Michoacana Meat Market. The number of each Preparatory School District's T\&T Supermarket equals 4 more than each Engineering Campus's Seafood City Supermarket. The number of each Science Park's T\&T Supermarket equals 4. The number of each Engineering Campus's La Michoacana Meat Market equals 0. The number of each Engineering Campus's T\&T Supermarket equals 1 times as much as the difference of each Engineering Campus's La Michoacana Meat Market and each Preparatory School District's Seafood City Supermarket. The number of each Engineering Campus's Seafood City Supermarket equals 2 times as much as the sum of each Science Park's Seafood City Supermarket, each Science Park's La Michoacana Meat Market and each Science Park's T\&T Supermarket. The number of each Science Park's La Michoacana Meat Market equals 3 times as much as each Science Park's T\&T Supermarket. The number of each Preparatory School District's La Michoacana Meat Market equals 3 more than the difference of each Science Park's T\&T Supermarket and each Science Park's La Michoacana Meat Market. \textbf{The number of each Science Park's Zion Market equals 1 more than each Arts Campus's T\&T Supermarket.} The number of each Engineering Campus's Zion Market equals each Engineering Campus's T\&T Supermarket. \textbf{How many Zion Market does Science Park have?}}\\

$\blacktriangleright$ \textbf{Output:}\\
Define Arts Campus's T\&T Supermarket as e; so e = 3. Define Science Park's Zion Market as w; so w = e + 1 = 3 + 1 = 4.
\end{tcolorbox}

\begin{tcolorbox}[
    breakable,
    colframe=black!75,
    colback=gray!10,
    coltitle=white,
    colbacktitle=black!80,
    title=\textbf{Non-IC Sample (Operations = 2)},
    fontupper=\small,
    fonttitle=\bfseries
]
$\blacktriangleright$ \textbf{Input:}\\
\texttt{\textbf{The number of each Arts Campus's T\&T Supermarket equals 3. The number of each Science Park's Zion Market equals 1 more than each Arts Campus's T\&T Supermarket.} \textbf{How many Zion Market does Science Park have?}}\\

$\blacktriangleright$ \textbf{Output:}\\
Define Arts Campus's T\&T Supermarket as e; so e = 3. Define Science Park's Zion Market as w; so w = e + 1 = 3 + 1 = 4.
\end{tcolorbox}

\subsection{Testing on Closed-Sourced Model}
\label{test_close_source}
To evaluate closed-source models, we use GPT-4o-mini to test across all operations. Additionally, we included a Background from the underlying graph structure to explicitly tell the model entity relationships, helping the model construct the correct reasoning context. Since the model struggles to learn modular operations, we also embed five-shot prompting.
\label{appendix:prompt}

\begin{tcolorbox}[
    breakable,
    colframe=black!75,
    colback=gray!10,
    coltitle=white,
    colbacktitle=black!80,
    title=\textbf{5-shots Testing Sample(Operations = 2)},
    fontupper=\small,
    fonttitle=\bfseries
]
$\bigstar$ \textbf{System:}\\
\texttt{You're an expert at solving elementary math problems involving addition, subtraction, and multiplication. You solve all the problems in a uniform format. All calculations are done modulo 5. For example, 3 + 2 equals 0, 1 + 1 equals 2, 4 + 2 + 4 equals 0, 3 * 2 equals 1, and 3 * 1 equals 3. When providing your solution, please end with 'The final answer is <<x>>.' where x is your final answer, an integer between 0 and 4. You must solve all the problems using the same solution format. Our scenarios involve up to four categories of objects:schools, classrooms, backpacks and stationeries. Each school may contain classrooms, each classroom may contain backpacks, and each backpack may contain stationeries. We can specify quantities, such as "the number of dance studios at each Lakeshore High." \\\\
Assume that every entity with the same name has an identical configuration; for example, each Lakeshore High
contains the same number of dance studios. Another guiding principle is that what is not mentioned does not
exist: when we refer to classrooms at Lakeshore High, we are only discussing the classrooms explicitly mentioned in our scenario. Furthermore, if Lakeshore High is not even mentioned, any classroom within it is automatically
considered to be non-existent (i.e. 0).}\\

$\blacktriangleright$ \textbf{User:} ...\\

$\blacktriangleright$ \textbf{Assistant:} ...\\
\\
$\blacktriangleright$ \textbf{User:} ...\\
$\blacktriangleright$ \textbf{Assistant:} ...\\
\\
$\blacktriangleright$ \textbf{User:} ...\\
$\blacktriangleright$ \textbf{Assistant:} ...\\
\\
$\blacktriangleright$ \textbf{User:} ...\\
$\blacktriangleright$ \textbf{Assistant:} ...\\
\\
$\blacktriangleright$ \textbf{User:} ...\\
$\blacktriangleright$ \textbf{Assistant:} ...\\

$\blacktriangleright$ \textbf{User:} \\
\textbf{Background: }\\
\texttt{There are 4 types of Zoo: Jurong Bird Park, Flamingo Gardens, Tracy Aviary, and Avery Island. There are 4 types of Enclosure: Ladybug Loft, Dragonfly Delta, Snail Shellter, and Beetle Bungalow. There are 2 types of Animal: Fire Salamander, and Newt. There are 3 types of Bone: Tertials, Secondary Feathers, and Metacarpals. Each Ladybug Loft's Fire Salamander can have Ladybug Loft's Animal. Each Tracy Aviary's Snail Shellter can have Snail Shellter's Newt and Ladybug Loft's Fire Salamander. Each Snail Shellter's Newt can have Ladybug Loft's Fire Salamander. Each Jurong Bird Park's Zoo can have Tracy Aviary's Snail Shellter, Snail Shellter's Newt, and Ladybug Loft's Fire Salamander. \\}
\\
\textbf{The problem description is: }\\
\texttt{The number of each Snail Shellter's Newt equals 4 more than each Tracy Aviary's Snail Shellter. The number of each Ladybug Loft's Fire Salamander equals 1 times as much as the difference of each Snail Shellter's Newt and each Tracy Aviary's Snail Shellter. The number of each Tracy Aviary's Snail Shellter equals 4. How many Animal does Ladybug Loft have?\\}

$\blacktriangleright$ \textbf{GPT-4o-mini Predicted Solution: \textcolor{red}{(Incorrect)}}\\
Define Tracy Aviary's Snail Shellter as T; so T = 4. \\
Define Snail Shellter's Newt as N; so N = T + 4 = 4 + 4 = 3. \\
Define Ladybug Loft's Fire Salamander as F; so F = N - T = 3 - 4 = 0. \textcolor{red}{\textbf{[CALCULATION ERROR]}}\\
Define Ladybug Loft's Animal as A; so A = F = 0.\\
\\
$\blacktriangleright$ \textbf{Groud-Truth Solution: \textcolor{green!60!black}{(Correct)}}\\
Define Tracy Aviary's Snail Shellter as o; so o = 4. \\
Define Snail Shellter's Newt as S; so S = 4 + o = 4 + 4 = 3. \\
Define Ladybug Loft's Fire Salamander as s; m = S - o = 3 - 4 = 4; so s = 1 * m = 1 * 4 = 4. \\
Define Ladybug Loft's Animal as H; so H = s = 4.\\
\\
$\bullet$ \textbf{Step Accuracy:} \textcolor{red}{False} \\
$\bullet$ \textbf{Path Accuracy:} \textcolor{green!60!black}{True} \\
$\bullet$ \textbf{Final Answer Accuracy:} \textcolor{red}{False}\\
\\
$\blacktriangleleft$ \textbf{Failure Reason:} \\ 
The model correctly selects every relevant entity and follows the intended dependency chain---first computing the Newt count \(N\) from the Snail Shellter count \(T\), then deriving the Fire Salamander count \(F\) from \(N\) and \(T\), and finally mapping \(F\) to the total Animals---showing no influence from irrelevant context \textcolor{green!60!black}{(Path Accuracy = True)}.  
Nonetheless, it commits a modular–arithmetic error: it evaluates \(F = N - T = 3 - 4\) as \(0\) instead of the correct value \(4\) under modulo 5.\textcolor{red}{(Step Accuracy = False, Final Answer Accuracy = False)}.
\end{tcolorbox}

\section{Quantifying Irrelevant Information}
\label{appendix:quan_irr_nodes}
To empirically study the impact of irrelevant information, we control the number of extraneous nodes and edges injected into each example (see \autoref{tab:irr_nodes}). These irrelevant parameters are randomly sampled from unused entities in the underlying graph, ensuring they do not alter the correct reasoning path. We incrementally adjust the number of injected nodes based on both model performance and problem difficulty. Notably, when the number of irrelevant nodes becomes large, model performance drops significantly. To avoid saturating the model's capacity and distorting evaluation, we refrain from injecting more irrelevant information beyond this point.
\begin{table}[H]
\centering
\footnotesize
\label{tab:extraneous_nodes}
\begin{tabular}{llccc}
\toprule
& \multirow{2}{*}{Operation} & \multicolumn{3}{c}{Irrelevant Parameters} \\
\cmidrule(lr){3-5}
& & Light & Medium & Hard \\
\midrule
& $op = 2$  & 0–2  & 3–4  & 5- \\
& $op = 3$  & 0–1  & 2–4  & 5– \\
& $op = 4$  & 0–1  & 2–3  & 4– \\
& $op = 5$  & 0–1  & 2–3  & 4– \\
& $op = 6$  & 0–1  & 2–3  & 4– \\
& $op = 7$  & 0–1  & 2–3  & 4– \\
& $op = 8$  & 0–1  & 2–3  & 4– \\
& $op = 9$  & 0–1  & 2–2  & 3– \\
& $op = 10$ & 0–1  & 2–2  & 3– \\
& $op = 11$ & 0–0  & 1–2  & 3– \\
& $op = 12$ & 0–0  & 1–2  & 3– \\
& $op = 13$ & 0–0  & 1–2  & 3– \\
& $op = 14$ & 0–0  & 1–2  & 3– \\
& $op = 15$ & 0–0  & 1–2  & 3– \\
& $op = 16$ & 0–0  & 1–1  & 2– \\
& $op = 17$ & 0–0  & 1–1  & 2– \\
& $op = 18$ & 0–0  & 1–1  & 2– \\
& $op = 19$ & 0–0  & 1–1  & 2– \\
& $op = 20$ & 0–0  & 1–1  & 2– \\
& $op = 21$ & 0–0  & 1–1  & 2– \\
\bottomrule
\end{tabular}
\caption{Quantile distribution of extraneous nodes across different operations.}
\label{tab:irr_nodes}
\end{table}

\label{appendix:synthetic}

\section{Clarifying the Role of Synthetic Problem Generation}

We clarify that the primary goal of our benchmark is to evaluate the logical reasoning capabilities of large language models (LLMs) under controlled conditions rather than to mimic natural human language generation. Many prior works similarly employ synthetic symbolic inputs to isolate and test reasoning abilities without the confounding influence of linguistic variability, including PrOntoQA~\cite{boratko-etal-2020-protoqa}, ProofWriter~\cite{tafjord2021proofwritergeneratingimplicationsproofs}, and iGSM~\cite{DBLP:journals/corr/abs-2407-20311}.

Although our directed acyclic graphs (DAGs) are randomly generated, the corresponding math word problems are constructed using four manually curated semantic hierarchies, each containing approximately 100 real-world entities per level. This design ensures that the resulting problems are semantically coherent and interpretable (\textit{e.g.}, ``A zoo’s enclosure has 3 giraffes’’ rather than ``A giraffe’s zoo has 3 enclosures’’). As such, our problems are more grounded than purely symbolic ones (\textit{e.g.}, ``$A + B = ?$’’) that lack real-world context.

Moreover, real-world math problems often require a combination of commonsense reasoning and complex arithmetic manipulation. For instance, commonsense tasks may involve knowledge composition (e.g., a chick has two legs and a rabbit has four legs)~\cite{allenzhu2024physicslanguagemodels32}, while arithmetic reasoning may require digit-level operations. Many studies have focused on these dimensions individually in synthetic environments. Our benchmark contributes to this space by isolating the effect of irrelevant context on logical reasoning—a critical step toward building robust and reliable LLMs.

\section{Operation‑Range Bias in \textsc{GSM‑IC}}
\label{appendix:gsm_ic}

\begin{figure}[t]
    \centering
    \includegraphics[width=0.9\linewidth]{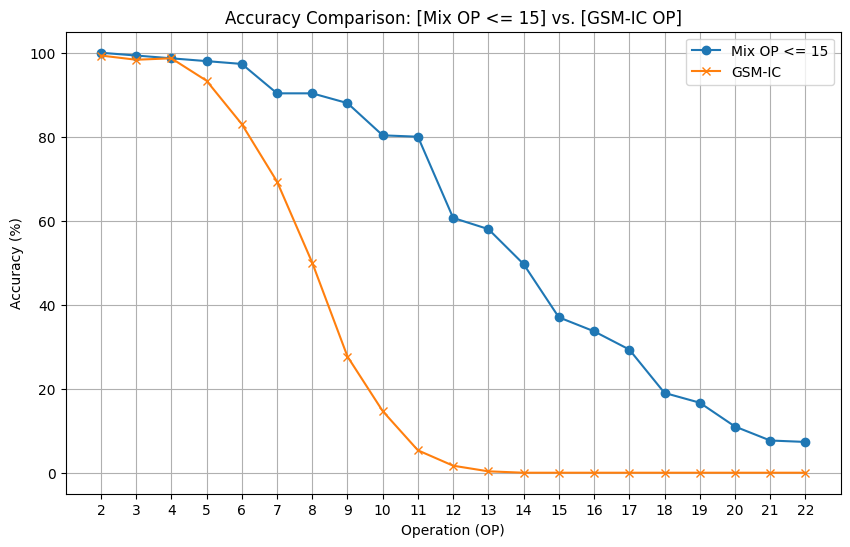}
    \caption{Test accuracy of a model trained on the \textsc{Mix $\leq15$} set (operation distribution identical to \textsc{GSM‑IC}) when evaluated on OP $=2$–$22$.  Accuracy drops substantially once the required operation count exceeds the training ceiling.}
    \label{img:gsmic}
\end{figure}

We found that models trained on problems containing only a small number of arithmetic operations tend to overfit short reasoning templates and fail to extrapolate to longer chains of computation.  To make this limitation explicit, we \textbf{adopt exactly the same operation‑count distribution as \textsc{GSM‑IC}} for all in‑distribution (ID) training examples (OP $=2$–$15$).  Generalisation is then probed with a held‑out out‑of‑distribution (OOD) slice comprising problems that require sixteen to twenty‑two operations.  \autoref{img:gsmic} plots test accuracy against operation count: performance remains high within the ID range but deteriorates rapidly once the task exceeds the training horizon, underscoring the necessity of our two‑tier protocol for a fair assessment of compositional reasoning.

\section{Process Reward Model}
\label{appendix:prm}

The Process Reward Model (PRM) we used in our experiments was trained using responses from an already fine-tuned model on problems of varying operation levels. Additionally, we utilized a parser that would identify exactly at which step of the LLM's reasoning chain the first mistake was made. We segmented the model's responses by using "." and ";" as our stop tokens, and labeled each segment depending on whether it is correct or not as illustrated below. Wherever the parser identified an error, that step and all subsequent steps would receive a negative label \textcolor{red}{\textbf{[$-$]}}, while all steps prior received a positive label \textcolor{green!60!black}{\textbf{[$+$]}}. 
\begin{tcolorbox}[
    breakable,
    colframe=black!75,
    colback=gray!10,
    coltitle=white,
    colbacktitle=black!80,
    title=\textbf{PRM Example with Correct Steps},
    fontupper=\small,
    fonttitle=\bfseries
]
$\blacktriangleright$ \textbf{Problem:}\\
The number of each Coniferous Forest's Hippopotamus equals 2 more than the difference of each Hippopotamus's Aortic Valve and each Hippopotamus's Organs. The number of each Tropical Dry Forest's Rhinoceros equals 2 more than the sum of each Coniferous Forest's Organs, each Rhinoceros's Organs and each Hippopotamus's Cardiac Muscle. The number of each Hippopotamus's Aortic Valve equals 1. The number of each Hippopotamus's Cardiac Muscle equals 3 times as much as each Hippopotamus's Aortic Valve. How many Organs does Tropical Dry Forest have?\\

$\blacktriangleright$ \textbf{Responses + Labels:}\\
    "Define Hippopotamus's Aortic Valve as g;" \textcolor{green!60!black}{\textbf{[$+$]}} \\
    "so g = 1." \textcolor{green!60!black}{\textbf{[$+$]}} \\
    "Define Hippopotamus's Cardiac Muscle as l;" \textcolor{green!60!black}{\textbf{[$+$]}} \\
    "so l = 3 * g = 3 * 1 = 3." \textcolor{green!60!black}{\textbf{[$+$]}} \\
    "Define Hippopotamus's Organs as G;" \textcolor{green!60!black}{\textbf{[$+$]}} \\
    "so G = g + l = 1 + 3 = 4." \textcolor{green!60!black}{\textbf{[$+$]}} \\
    "Define Coniferous Forest's Hippopotamus as y;" \textcolor{green!60!black}{\textbf{[$+$]}} \\
    "w = g - G = 1 - 4 = 2;', 'so y = 2 + w = 2 + 2 = 4." \textcolor{green!60!black}{\textbf{[$+$]}} \\
    "Define Coniferous Forest's Organs as L;" \textcolor{green!60!black}{\textbf{[$+$]}} \\
    "so L = y * G = 4 * 4 = 1." \textcolor{green!60!black}{\textbf{[$+$]}} \\
    "Define Rhinoceros's Organs as u;" \textcolor{green!60!black}{\textbf{[$+$]}} \\
    "so u = 0." \textcolor{green!60!black}{\textbf{[$+$]}} \\
    "Define Tropical Dry Forest's Rhinoceros as R;" \textcolor{green!60!black}{\textbf{[$+$]}} \\
    "s = L + u = 1 + 0 = 1;" \textcolor{green!60!black}{\textbf{[$+$]}} \\
    "C = s + l = 1 + 3 = 4;" \textcolor{green!60!black}{\textbf{[$+$]}} \\
    "so R = 2 + C = 2 + 4 = 1." \textcolor{green!60!black}{\textbf{[$+$]}} \\
    "Define Tropical Dry Forest's Organs as V;" \textcolor{green!60!black}{\textbf{[$+$]}} \\
    "so V = R * u = 1 * 0 = 0." \textcolor{green!60!black}{\textbf{[$+$]}}
\end{tcolorbox}


The use of a synthetic dataset offers two key advantages: it simplifies the curation of training data for our Process Reward Model (PRM) and enables precise control over injected mistakes, allowing us to label incorrect reasoning steps explicitly. This was made possible by a solution parser capable of not only verifying the final answer but also pinpointing the exact step where an error occurred. As a result, we were able to construct high-quality, fine-grained supervision signals to effectively train the PRM.

Furthermore, we are able to purposely inject IC into a given problem due to the synthetic nature of our dataset. For our experiments, after having constructed the directed graph of the correct problem, we add IC by generating extraneous details and directly including them within the given problem statement at various steps. We then pass this new problem into the LLM to get its response. Afterwards, we pass the LLM's response into our parser and it can identify the presence of and type of error that was made. Examples of errors include, but are not limited to, arithmetic errors, definitions of irrelevant/non-existent symbols, or duplicate symbols.

\begin{tcolorbox}[
    breakable,
    colframe=black!75,
    colback=gray!10,
    coltitle=white,
    colbacktitle=black!80,
    title=\textbf{PRM Example with Wrong Steps},
    fontupper=\small,
    fonttitle=\bfseries
]
$\blacktriangleright$ \textbf{Problem:}\\
The number of each Nasal Cavity's Pericytes equals the difference of each Nasal Cavity's Arrector Pili Muscle Cells and each Vocal Cords's Arrector Pili Muscle Cells. The number of each Nasal Cavity's Gastrointestinal Smooth Muscle Cells equals 3 times as much as each Vocal Cords's Arrector Pili Muscle Cells. The number of each Respiratory Mucosa's Pericytes equals 4 times as much as the sum of each Nasal Cavity's Gastrointestinal Smooth Muscle Cells and each Vocal Cords's Arrector Pili Muscle Cells. The number of each Vocal Cords's Pericytes equals 3. The number of each Respiratory Mucosa's Arrector Pili Muscle Cells equals 2. The number of each Nasal Cavity's Arrector Pili Muscle Cells equals 3 more than the difference of each Nasal Cavity's Gastrointestinal Smooth Muscle Cells and each Respiratory Mucosa's Pericytes. The number of each Vocal Cords's Arrector Pili Muscle Cells equals 3 times as much as each Vocal Cords's Gastrointestinal Smooth Muscle Cells. The number of each Respiratory Mucosa's Gastrointestinal Smooth Muscle Cells equals each Nasal Cavity's Arrector Pili Muscle Cells. The number of each Vocal Cords's Gastrointestinal Smooth Muscle Cells equals 4. How many Cells does Respiratory Mucosa have?\\

$\blacktriangleright$ \textbf{Responses + Labels:}\\
    "Define Vocal Cords's Gastrointestinal Smooth Muscle Cells as W;" \textcolor{green!60!black}{\textbf{[$+$]}} \\
    "so W = 4." \textcolor{green!60!black}{\textbf{[$+$]}} \\
    "Define Vocal Cords's Arrector Pili Muscle Cells as p;" \textcolor{red}{\textbf{[$-$]}} \\
    "so p = 3 * W = 3 * 4 = 2." \textcolor{red}{\textbf{[$-$]}} \\
    "Define Nasal Cavity's Gastrointestinal Smooth Muscle Cells as g;" \textcolor{red}{\textbf{[$-$]}} \\
    "so g = 3 * p = 3 * 2 = 1." \textcolor{red}{\textbf{[$-$]}} \\
    "Define Nasal Cavity's Arrector Pili Muscle Cells as e;" \textcolor{red}{\textbf{[$-$]}} \\
    "c = g - p = 1 - 2 = 4;" \textcolor{red}{\textbf{[$-$]}} \\
    "so e = 3 + c = 3 + 4 = 2." \textcolor{red}{\textbf{[$-$]}} \\
    "Define Respiratory Mucosa's Pericytes as z;" \textcolor{red}{\textbf{[$-$]}} \\
    "w = g + p = 1 + 2 = 3;" \textcolor{red}{\textbf{[$-$]}} \\
    "so z = 4 * w = 4 * 3 = 2." \textcolor{red}{\textbf{[$-$]}} \\
    "Define Respiratory Mucosa's Arrector Pili Muscle Cells as F;" \textcolor{red}{\textbf{[$-$]}} \\
    "so F = 2." \textcolor{red}{\textbf{[$-$]}} \\
    "Define Respiratory Mucosa's Gastrointestinal Smooth Muscle Cells as P;"" \textcolor{red}{\textbf{[$-$]}} \\
    so P = e = 2." \textcolor{red}{\textbf{[$-$]}} \\
    "Define Respiratory Mucosa's Cells as m;" \textcolor{red}{\textbf{[$-$]}} \\
    "G = P + F = 2 + 2 = 4;" \textcolor{red}{\textbf{[$-$]}} \\
    "so m = G + z = 4 + 2 = 1." \textcolor{red}{\textbf{[$-$]}} \\
$\blacktriangleright$ \textbf{Parser:}\\
    existing\_but\_not\_required\_params: Vocal Cords's Arrector Pili Muscle Cells
\end{tcolorbox}

Thus, this allows us to control the amount of IC present and ultimately measure the effects of unnecessary information on LLM's responses.
We prepared a dataset of 5000 problems of varying OP values $\in$ [2, 15] as well as an additional 1000 problems of OP=15 so that the model had a sufficient number of high operation training problems. 
Ultimately, the PRM was trained on each problem and each of its steps and used to facilitate our Tree of Thoughts (ToT) algorithm. 

In settings with a PRM, we generated responses in a step by step manner by using ";" and "." as our intermediary stop tokens. Each intermediary step would be scored by the PRM and only the top $N / M$ responses would be selected as candidates in the next step to be explored further. This process was repeated until the LLM generated the \textit{<EOS>} token, signaling that the response was complete. This final response would then be passed into the parser to determine its correctness.

\section{Additional Analyses}
\label{appendix:additional}

\paragraph{Disentangling Reasoning Steps (\texorpdfstring{$rs$}{rs}) from Irrelevant Context (IC)}
\label{appendix:c3}

Our primary objective is to understand how irrelevant context (IC) affects LLM reasoning. Hence, our main experiments fix the reasoning-step depth ($rs$) and systematically manipulate IC intensity. For completeness, we report the IC = 0 baselines across increasing $rs$ for two models (\textsc{Llama-3.2-1B-Instruct} and \textsc{Llama-3.3-70B-Instruct}) and two reasoning steps $(rs)$ settings ($rs$ = 2 and $rs$ = 4) in \autoref{fig:ic0-vertical}. These results do not change our main conclusions (\textit{e.g.}. Result II); they confirm that deeper reasoning paths reduce accuracy even without distractors.

\begin{figure}[t]
\centering
\pgfplotsset{
  ic0plot/.style={
    width=\columnwidth,
    height=3.3cm,
    xmin=0, xmax=15,
    ymin=0, ymax=100,
    xtick={0,1,3,5,7,9,11,13,15},
    ytick={0,20,40,60,80,100},
    grid=both,
    tick label style={font=\scriptsize},
    label style={font=\scriptsize},
    title style={font=\footnotesize, yshift=-4pt},
    legend style={font=\scriptsize},
    cycle list={
      {metricS, mark=*, mark size=1.8pt, thick},
      {metricP, mark=square*, mark size=1.6pt, thick},
      {metricE, mark=triangle*, mark size=2pt, thick}
    }
  }
}
\begin{tikzpicture}
\begin{groupplot}[
  group style={group size=1 by 4, vertical sep=30pt},
  ic0plot,
  ylabel={Accuracy (\%)},
]

\nextgroupplot[
  title={\textsc{LLaMA-3.2-1B-Instruct}, $rs$ = 2},
  xticklabels=\empty, xlabel={}
]
\addplot coordinates {(0,41) (1,9) (3,8) (5,3) (7,4) (9,4) (11,1) (13,1) (15,2)};
\addplot coordinates {(0,44) (1,14) (3,12) (5,10) (7,8) (9,8) (11,5) (13,3) (15,4)};
\addplot coordinates {(0,72) (1,56) (3,48) (5,41) (7,36) (9,39) (11,28) (13,45) (15,38)};
\legend{SAcc, PAcc, EAcc}

\nextgroupplot[
  title={\textsc{LLaMA-3.2-1B-Instruct}, $rs$ = 4},
  xticklabels=\empty, xlabel={}
]
\addplot coordinates {(0,3) (1,1) (3,0) (5,0) (7,0) (9,0) (11,0) (13,0) (15,0)};
\addplot coordinates {(0,3) (1,2) (3,0) (5,0) (7,0) (9,0) (11,1) (13,0) (15,0)};
\addplot coordinates {(0,68) (1,49) (3,45) (5,41) (7,35) (9,29) (11,32) (13,29) (15,27)};

\nextgroupplot[
  title={\textsc{LLaMA-3.3-70B-Instruct}, $rs$ = 2},
  xticklabels=\empty, xlabel={}
]
\addplot coordinates {(0,76) (1,53) (3,49) (5,59) (7,51) (9,51) (11,59) (13,48) (15,52)};
\addplot coordinates {(0,76) (1,53) (3,49) (5,59) (7,51) (9,51) (11,59) (13,49) (15,52)};
\addplot coordinates {(0,98) (1,98) (3,89) (5,95) (7,94) (9,95) (11,88) (13,90) (15,87)};

\nextgroupplot[
  title={\textsc{LLaMA-3.3-70B-Instruct}, $rs$ = 4},
  xlabel={$rs$}
]
\addplot coordinates {(0,61) (1,27) (3,29) (5,24) (7,26) (9,29) (11,26) (13,20) (15,14)};
\addplot coordinates {(0,63) (1,33) (3,29) (5,24) (7,26) (9,29) (11,26) (13,20) (15,15)};
\addplot coordinates {(0,97) (1,76) (3,79) (5,76) (7,57) (9,74) (11,66) (13,66) (15,44)};

\end{groupplot}
\end{tikzpicture}

\caption{IC = 0 baselines across reasoning steps ($rs$).}
\label{fig:ic0-vertical}
\end{figure}

\paragraph{Reporting Final Answer Accuracy (EAcc)}
\label{appendix:w2a}

While we computed EAcc throughout, our main paper emphasizes SAcc and PAcc because they directly evaluate whether the model follows a correct reasoning chain under distraction. Below we provide the full EAcc results; the overall conclusions are unchanged when using EAcc.

\begin{table}[t]
\centering
\setlength{\tabcolsep}{4pt}
\footnotesize
\resizebox{\columnwidth}{!}{%
\begin{tabular}{lccc|ccc}
\toprule
\multirow{2}{*}{\makecell{Training\\IC Level}} 
  & \multicolumn{3}{c|}{\textbf{Testing w/ IC (EAcc)}} 
  & \multicolumn{3}{c}{\textbf{Testing w/o IC (EAcc)}} \\
\cmidrule(lr){2-4}\cmidrule(lr){5-7}
 & ID & OOD & All & ID & OOD & All \\
\midrule
\textsc{Clean}     & 68.64 & 53.29 & 63.52 & 90.07 & 57.38 & 79.13 \\
\textsc{Light-IC}  & 80.66 & 44.08 & 68.46 & 86.87 & 45.09 & 72.94 \\
\textsc{Medium-IC} & 82.69 & 47.91 & 71.09 & 85.84 & 49.14 & 73.60 \\
\textsc{Hard-IC}   & \textbf{89.75} & \textbf{58.06} & \textbf{79.19} & \textbf{92.86} & \textbf{62.84} & \textbf{82.85} \\
\textsc{Mix-IC}    & 87.26 & 54.81 & 76.44 & 90.91 & 55.41 & 79.08 \\
\bottomrule
\end{tabular}%
}
\caption{EAcc under different training IC levels, with/without IC at test time. The ordering mirrors SAcc/PAcc trends, reinforcing our main claims.}
\label{tab:w2a-eacc-ic}
\end{table}

\noindent In short, EAcc corroborates the overall picture but may overestimate robustness when models land on the correct final answer via an incorrect or IC-contaminated chain. This is why we emphasize SAcc and PAcc for evaluating logical reasoning under distraction.
\section{Finetuning Details}
\label{appendix:gpu}

\paragraph{Model}
We finetune \textbf{LLaMA 3.2-1B Instruct} model released by Meta using both LoRA finetuning and continued pretraining (full finetuning). This model adopts a decoder-only transformer architecture with rotary positional embeddings and gated MLP layers. All experiments are performed on two NVIDIA H100 GPUs.

\paragraph{Finetuning Configuration}
Due to some computational constraints, our training is conducted on a fixed dataset of 30{,}000 samples. Each example contains a complete problem-solution pair, and inputs exceeding 2048 tokens are filtered out. We use a context length of 2048, a learning rate of 5e\textminus5, and the AdamW optimizer with cosine learning rate decay. Training proceeds for 50 epochs with a batch size of 8 and gradient accumulation of 8 steps, yielding an effective batch size of 64. We apply mixed-precision training with \texttt{bfloat16}, no warmup, and a maximum gradient norm of 1.0. Flash attention is enabled.

\paragraph{Evaluation Protocol}
We evaluate each model on a fixed test set containing 100 examples per reasoning operation and per level of irrelevant context (IC), including \emph{Light}, \emph{Medium}, and \emph{Hard}. Evaluations are performed separately on in-distribution (ID) and out-of-distribution (OOD) data. This setup enables precise measurement of reasoning robustness under varying levels of distractibility, supporting our core analysis of how irrelevant information affects model behavior.

\paragraph{Architectural Generalization.}
Recent controlled studies suggest that decoder-only transformer models equipped with full attention and rotary positional embeddings tend to exhibit similar learning dynamics and inductive biases, even when implemented under different architectures. These models—such as GPT-style, LLaMA, Mixtral, and others—differ in details like normalization placement or gated MLPs, but such variations do not appear to fundamentally alter their learnability or reasoning behavior in practice \citep{allenzhu2024physicslanguagemodels1,allenzhu2024physicslanguagemodels32,allenzhu2024physicslanguagemodels33}. In our case, although early experiments were conducted using a LLaMA-style architecture, all final results presented in this paper are based on the more recent \textit{LLaMA 3.2–1B Instruct} model. We did not observe substantial performance differences across architectures during preliminary runs. Given resource constraints, we focus on LLaMA 3.2–1B in this version; however, we acknowledge that running a comprehensive comparison across reasoning models (\textit{e.g.} DeepSeek-R1) would strengthen the generality of our findings and plan to pursue this in future work.

\paragraph{Does More In-Distribution Data Help?}
\label{appendix:data_scaling}

\begin{figure}[ht]
    \centering
    \includegraphics[width=0.48\textwidth]{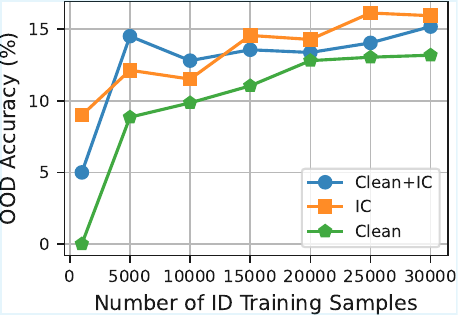}
  \caption{OOD Step Accuracy (SAcc) as a function of in-distribution training size.}
    \label{fig:id_samples_ood}
\end{figure}

To identify an effective training budget, we varied the number of in‑distribution samples from 1 K to 30 K and observed saturating OOD gains around 30 K (\autoref{fig:id_samples_ood}). Based on this, we fixed 30 K samples for all subsequent experiments.

\section{Future Work}
\label{appendix:future}
We will scale model capacity and context length to assess robustness across parameter regimes, and explore advanced parameter-efficient fine-tuning to improve the robustness–efficiency trade-off~\citep{chen2024longloraefficientfinetuninglongcontext,xu2023qaloraquantizationawarelowrankadaptation,huang2025loraparflexibledualsystemlora}. To broaden the generality of our PRM+ToT inference, we will retrain the PRM on diverse external reasoning corpora (e.g., ProofWriter, StrategyQA) and benchmark adaptive search policies. Finally, inspired by fine-grained reliability work in VLMs, we will extend our IC injectors and PRM scoring to multimodal settings and evaluate robustness on vision–language and multi-modal benchmarks~\citep{dong2025benchmarkingretrievalaugmentedmultimomalgeneration,wang2025reliablevlmfinegrainedbenchmark}.

\end{document}